\documentclass[conference]{IEEEtran}
\IEEEoverridecommandlockouts

\usepackage{hyperref}       
\usepackage{url}            
\usepackage{booktabs}       
\usepackage{amsfonts}       
\usepackage{nicefrac}       
\usepackage{microtype}      
\usepackage{xcolor}         
\usepackage{amsmath}

\usepackage{graphicx}
\usepackage{comment}
\usepackage{textcomp}
\usepackage{subcaption}
\usepackage{bm}
\usepackage{wrapfig}

\usepackage{mathtools,xparse}
\usepackage{multirow}
\usepackage{makecell}
\usepackage{pifont}
\newcommand{\cmark}{$\checkmark$}
\newcommand{\xmark}{$\times$}

\DeclarePairedDelimiter{\norm}{\lVert}{\rVert}

\AtBeginDocument{%
  \providecommand\BibTeX{{%
    \normalfont B\kern-0.5em{\scshape i\kern-0.25em b}\kern-0.8em\TeX}}}

\begin{document}

\title{Spiking Neural Networks with Dynamic Time Steps for Vision Transformers \\[-1.5ex]}


\author{\IEEEauthorblockN{Gourav Datta$^*$, Zeyu Liu$^*$, Anni Li, and Peter A. Beerel}
\thanks{$^*$Equally contributing authors.} 
\thanks{$^{\dagger}$This work was supported by a gift funding from Intel.}
\it{Ming Hsieh Dept. of Electrical and Computer Engineering, University of Southern California, Los Angeles, USA} \\
 	\it{\{gdatta, liuzeyu, annili, pabeerel\}@usc.edu} \\[-1.0ex]}

\maketitle



\begin{abstract}
Spiking Neural Networks (SNNs) have emerged as a popular spatio-temporal computing paradigm for complex vision tasks. Recently proposed SNN training algorithms have significantly reduced the number of time steps (down to 1) for improved latency and energy efficiency, however, they target only convolutional neural networks (CNN). 
These algorithms, when applied on the recently spotlighted vision transformers (ViT), either require a large number of time steps or fail to converge. Based on analysis of the histograms of the ANN and SNN activation maps, we hypothesize that each ViT block has a different sensitivity to the number of time steps. We propose a novel training framework that dynamically allocates the number of time steps to each ViT module depending on a trainable score assigned to each timestep. 
In particular, we generate a scalar binary time step mask that filters spikes emitted by each neuron in a leaky-integrate-and-fire (LIF) layer. 
The resulting SNNs have high activation sparsity and require only accumulate operations (AC), except for the input embedding layer, in contrast to expensive multiply-and-accumulates (MAC) needed in traditional ViTs. This yields significant improvements in energy efficiency. We evaluate our training framework and resulting SNNs on image recognition tasks including CIFAR10, CIFAR100, and ImageNet with different ViT architectures. We obtain a test accuracy of 95.97\% with 4.97 time steps with direct encoding on CIFAR10. 
\end{abstract}






\section{Introduction}

Spiking neural networks (SNN) have  emerged as an attractive edge-based computing paradigm, thanks to their high activation sparsity and use of cheap accumulation (AC) instead of energy-expensive multiply-and-accumulation (MAC) \cite{spike_ratecoding,dsnn_conversion1}. Early research on SNNs suggested that they can achieve good performance in a wide range of modern vision tasks \cite{dsnn_conversion1}, but not without a noticeable accuracy drop compared to the state-of-the-art (SOTA), particularly when restricted to energy-efficient low-latency SNNs. This motivated a plethora of research on supervised SNN training algorithms, that can generally be grouped into i) ANN-to-SNN conversion, and ii) direct training. ANN-to-SNN conversion leverages standard backpropagation for training in the ANN domain, and yields SNNs by approximating the non-linear activation functions (e.g. ReLU in CNNs) with the firing rate of spiking neurons \cite{dsnn_conversion_abhronilfin,dsnn_conversion1,deng2021optimal,datta2022date}. In contrast, direct training algorithms typically leverage surrogate gradient learning that approximates the discontinuous spiking neuron functionality with a continuous differentiable model 
\cite{Wu_Deng_Li_Zhu_Xie_Shi_2019,panda_res,datta_ijcnn, datta2022hoyer}. While direct training algorithms require the gradients to be integrated over all the SNN time steps, thereby increasing the training complexity, it requires far less number of time steps compared to ANN-to-SNN conversion. To obtain the best of both worlds, a few research works \cite{rathi2021diet,Rathi2020Enabling,datta_frontiers,datta_lstm} attempted to perform ANN-to-SNN conversion, followed by fine-tuning in the SNN domain.
The latest SNN literature \cite{deng2022temporal,chowdhury20221-step,li2023seenn} suggests that direct training is possible with only a few time steps ($1$ or $2$) and with a slight increase in training complexity compared to non-spiking networks. These works significantly decrease latency and energy of SNNs while not compromising accuracy, thereby enabling the deployment of SNNs at the extreme edge.


In spite of these tremendous improvements in SNN training algorithms, most (if not all) existing SNN models are based on CNN backbones, such as VGG \cite{vgg} and ResNet \cite{resnet}. 
On the other hand, the last two years have witnessed the dominance of vision transformers (ViT) \cite{dosovitskiy2021an} that have outperformed SOTA CNNs in complex vision tasks. There is little research on SNNs for ViTs, particularly for low time steps and edge applications, where SNNs are the most attractive candidates for edge efficiency.
Existing works propose several architectural modifications in ViTs for SNNs that incur a large number of time steps ($4$ and above) to obtain similar accuracy as the non-spiking counterparts \cite{spikformer,li2022spikeformer}. There has not been any thorough analysis on the spiking activation functions for ViTs (in particular the self-attention block that is unique to ViTs) or training optimizations that reduce the number of time steps. 

With this motivation, this paper first shows that the existing CNN-based low-latency SNN training approaches cannot be ported to ViTs, without a significant drop in accuracy, particularly at low time steps, implying there are large gaps in both the accuracy and efficiency advantages of the traditional vision and SNN models. In order to mitigate these gaps, we investigate the key sources of information loss in ViT-based SNNs and propose a novel training algorithm that allocates time steps dynamically to each ViT module based on its precision sensitivity. Our algorithm yields SNN models that can obtain close to SOTA accuracy with an average number of time steps as low as ${\sim}2.73$ on the CIFAR100 dataset.
\section{Preliminaries}

\subsection{LIF Model \& Surrogate gradient}

In this work, we adopt the popular leaky-integrate-and-fire (LIF) neuron \cite{leefin2020} models to capture the computation dynamics of an SNN, where each neuron in the $l^{th}$ layer has an internal state called its membrane potential $U_l^t$ ($t$ denotes the time step). It captures the sum of the weight $W_l$ modulated incoming spikes $O_{l{-}1}^t$ from each neuron in the previous layer $(l{-}1)$ as shown below. 
\begin{align}
H_l^t&=\lambda_l U_l^{t{-}1}+W_l O_{l{-}1}^t \ \ O_l^t &=
\begin{cases}
    V_l^{th}, & \text{if } H_l^t>V_l^{th}\label{eq:lif_output}\\
    0, & \text{otherwise} \\
\end{cases} \\ 
U_l^t &= H_l^t-O_l^t
\label{eq:IF_out_spike}
\end{align}
 Note that $U_l^t$ leaks with a fixed time constant, denoted as $\lambda_l$. 
Since the spiking neuron function $O_l^t$ is discontinuous and non-differentiable, it is difficult to implement gradient descent based backpropagation in SNNs. Hence, previous works \cite{panda_res,neftci_surg} approximate the spiking function with a continuous differentiable function, which helps back-propagate non-zero gradients known as surrogate gradients. 
The resulting weight update in the $l^{th}$ hidden layer in the SNN is calculated as  
\begin{align}
    \Delta{W_l}{=}\sum_{t}\frac{\partial\mathcal{L}}{\partial W_l}{=}\sum_{t}\frac{\partial\mathcal{L}}{\partial\mbox{$O$}_l^t}\frac{\partial\mbox{$O$}_l^t}{\partial\mbox{$H$}_l^t}\frac{\partial\mbox{$H$}_l^t}{\partial W_l}{=}\sum_{t}\frac{\partial\mathcal{L}}{\partial\mbox{$O$}_l^t}\frac{\partial\mbox{$O$}_l^t}{\partial\mbox{$H$}_l^t}\mbox{$O$}_{l-1}^t \notag
\end{align}
\noexpand
where $\frac{\partial{O}_l^t}{\partial{U}_l^t}$ is the non-differentiable gradient which can be approximated with the surrogate gradient $\frac{\gamma}{V_l^{th}}\cdot{max(0,1-\text{abs}(\mbox{$\frac{U_l^t}{V_l^{th}}-1$))}}$, where 
$\gamma$ is a hyperparameter denoting the maximum gradient value.

\subsection{Architectural Modifications for Spiking ViT}

Inspired by \cite{spikformer}, we adopt some architectural modifications to the standard ViT backbone to simplify the training in the SNN domain. Firstly, we remove the Dropout and Droppath present in standard ViTs. Secondly, we remove the layer norm before each self-attention and multi-layer perceptron (MLP) block, and instead, add batch norm (BN) after each linear layer. Thirdly, we remove the softmax layer in the multi-head self-attention (MSA) block since the attention maps computed from the spiking query and key outputs are always positive. This is because we use a spiking LIF layer before every dot-product operation incurred in convolutional/linear/MSA layers so that we require only accumulate operations for energy-efficiency, as promised in SNNs. Details of our spiking ViT architecture are further illustrated in Fig. 1.

\section{Proposed Method}

Based on the LIF model, we propose a Dynamic Time Step Spiking layer (DTSS layer) that adaptively assigns different time steps to different LIF layers of the spiking ViT. Intuitively, this scheme can skip the computations and latency incurred in processing timesteps that can be potentially unnecessary in some layers. Furthermore, we analyzed the distribution of SNN activation maps for each layer of the Spikformer \cite{spikformer} at different timesteps. This analysis shows that the sensitivity of each transformer layer varies with the different time steps. Based on this intuition and analysis, we propose a dynamic time step spiking ViT training framework, that achieves close to SOTA test accuracy with lower latency.

\begin{figure*}[htbp]
\centering
\begin{subfigure}{0.48\textwidth}
    \includegraphics[width = \textwidth]{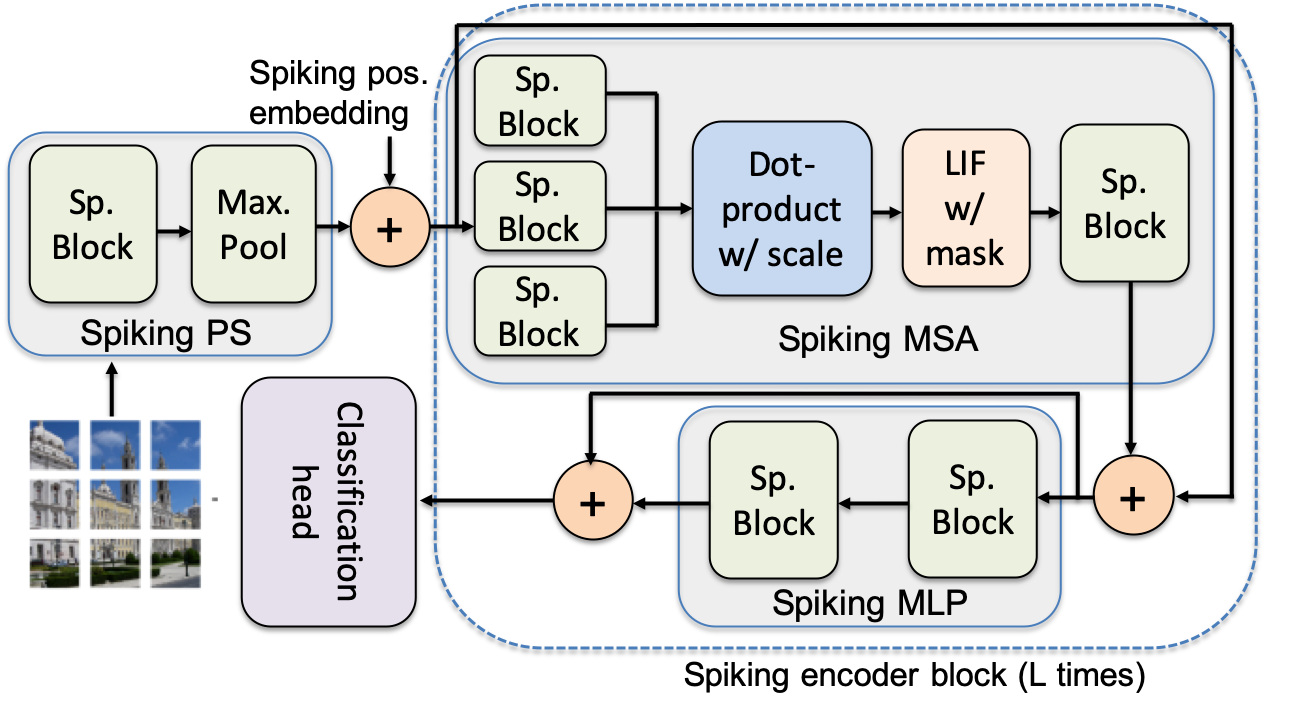}
    \caption{ }
    \label{fig:spikeformer}
    
\end{subfigure}
\begin{subfigure}{0.48\textwidth}
    \includegraphics[width = \textwidth]{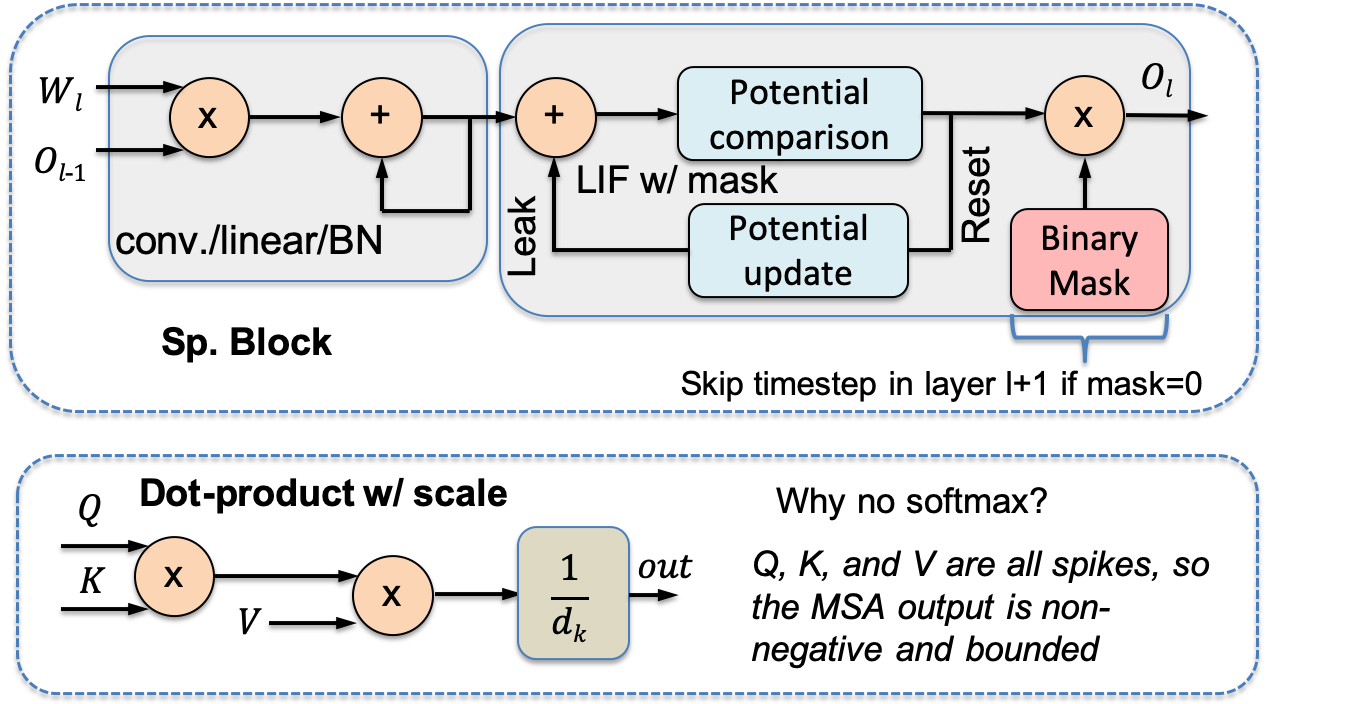}
    \caption{ }
    \label{fig:sp_block}
\end{subfigure}
\hfill
\caption{(a) Proposed network architecture of the spiking ViT used in this work, where PS denotes patch splitting and (b) Proposed dynamic time step spiking block and the MSA block (dot-product w/ scale) without the softmax layer.}
\label{fig:P2M_circuit_waves}
\vspace{-1mm}
\end{figure*}

\subsection{Dynamic Time Steps Spiking Layer}

In this subsection, we present our DTSS layer based on the traditional multi-time-step LIF model. Specifically, we first set a maximum number of time steps $T_{max}$ for our SNN model, which means that the number of time steps assigned to each layer is less than or equal to $T_{max}$. We then define a trainable tensor of length $T_{max}$, called the parameter tensor $TP_l$, and an all-one lower triangular matrix of size $T_{max}{\times}T_{max}$ called the coefficient matrix $\it{coeff}$. The parameter tensor $TP_l$ is initialized with all non-negative values, so that $TS_l$, which we denote as the score tensor, and which is the product of $TP_l$ and $\it{coeff}$ (see Eq. 5 below), is a non-increasing tensor. This helps ensure that the score tensor $TS_l$ represents a valid time step mask such that we do not skip any intermediate time steps i.e., we only skip time steps towards the end. 

To convert the score tensor at a particular time step $t$ ($TS_l^t$) to a binary mask tensor denoted by $TM_l^t$, we propose a simple spiking layer as illustrated in Eq.\eqref{eq:surrogate}. Since it is not possible to update $TP_l$ with the real gradients backpropagated from $TM_l^t$ (gradients are zero almost everywhere), we use surrogate gradients. 
Note that this spiking layer is different from the LIF layers used in the spiking ViT network architecture. As illustrated in Fig. 1, $TM_l^t$ dynamically retains the output of the LIF neurons of a particular layer in some specific time steps while setting the output in the remaining time steps to zero. This helps in shutting off the computations incurred in the subsequent layer for the remaining time steps, which increases the compute efficiency of the SNN. Moreover, during inference when $TM_l^t$ is fixed, this also reduces the number of time steps of the subsequent layer, which with appropriate spike routing support from the underlying hardware compiler, can also increase the SNN latency efficiency. To quantify this latency reduction, we define the total number of time steps of our SNN as
\begin{equation}\label{eq:overall:ts}
    T_{avg} = \frac{\sum_{l=1}^L T_{l}}{L}, l \in [1, 2, 3, ... , L]
\end{equation}
where $T_l$ is the total number of time steps of the $l^{th}$ layer, and $L$ is the total number of layers.

Without loss of generality, let us illustrate the function of our DTSS layer for the case when $T_{max}{=}4$. 
Denoting ${O}_l^t$ as the input and ${W}_l$ as the weights of the convolutional/linear layer $l$, we first obtain the temporary spiking outputs $Y_l^t$ following the LIF model illustrated in Eq. (1-3). Let us denote this model as $SN$.  
We then apply our mask tensor $TM_l^t$ to filter out the final spiking outputs ${O}_l^t$ that would be fed to layer $l{+}1$ only when all the elements of ${O}_l^t$ are not zero. 

\begin{equation}
\label{eq:coefficent}
\small{
    coeff{=} 
  \begin{bmatrix}
  1 & 0 & 0 & 0 \\
  1 & 1 & 0 & 0 \\
  1 & 1 & 1 & 0 \\
  1 & 1 & 1 & 1 
  \end{bmatrix}
  \; TS_l{=}TP_l{\cdot}coeff{=}[TS_l^0,TS_l^1,TS_l^2,TS_l^3]}
\end{equation}
\begin{equation} \label{eq:surrogate}
    {TM}_l^t = \Theta({TS}_l^t){=} \begin{cases} 
    1, & \text{if} {TS}_l^t \geq 1; \\ 
    0, & \text{otherwise} \end{cases}
\end{equation}
\begin{equation}
    \frac{\partial ({TM}_l^t)}{\partial ({TS}_l^t)}{=}
    \begin{cases} 
    1, & \text{if } 0 < {TS}_l^t < 2; \\ 
    0, & \text{otherwise} \end{cases}
\end{equation}

\begin{equation}\label{eq:DTS}
    {Y}_l^t  = SN({W}_l{O}_{l{-}1}^t)
    \; \; \; \;
    {O}^{t}_l = {Y}_l^t * TM_l^t
\end{equation}

\subsection{Distribution of SNN Membrane Potentials}
\label{sec:dist}

\begin{figure*}[htbp]
\centering
\includegraphics[width=1.0\textwidth]{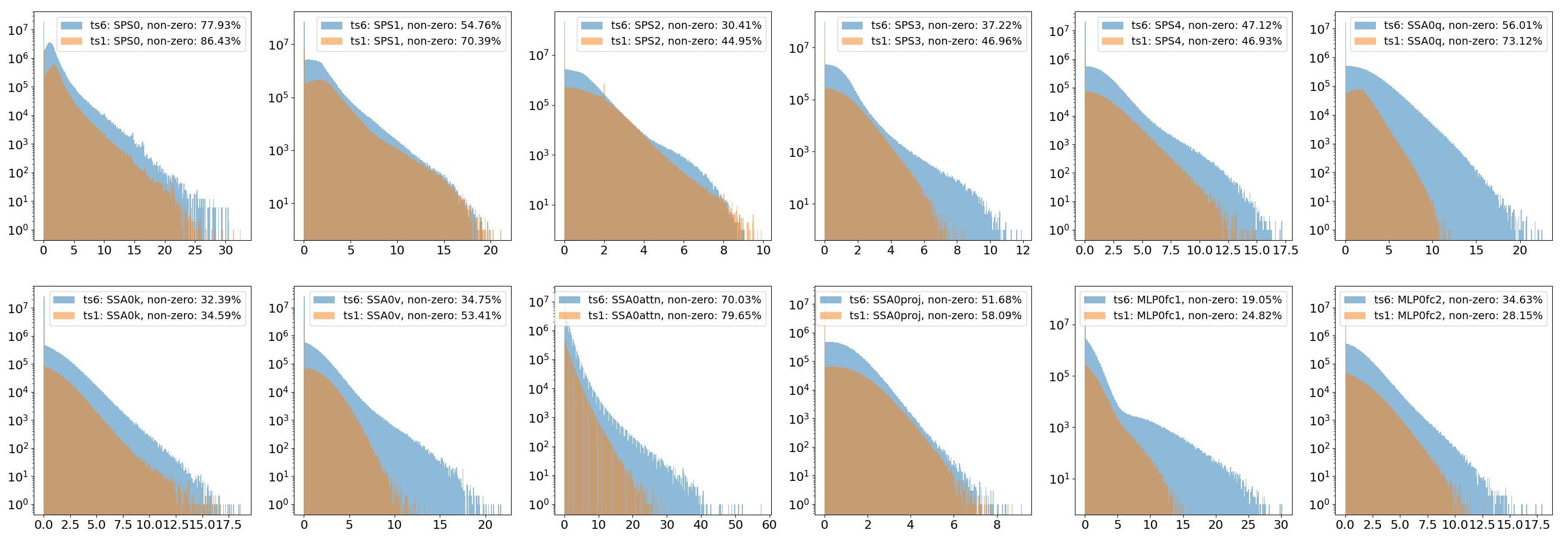}
\caption{The distribution of activation maps of the spiking models, the orange color means the distribution of the one-time-step model while the blue color means the distribution of the six-time-steps model. The label in each plot shows the name of the layer. 
The first five figures are the distributions of the layers in the spiking patch splitting (SPS) module, the next five are the distributions of the layers in the spiking self-attention (SSA) block, and the last two are the distributions of the layers in the multi-layer perception (MLP) block.
To keep it simplified, we omit the distributions of the remaining layers.}
\label{fig:dist:combine}
\end{figure*}

Apparently, the number of time steps in a spiking ViT layer $l$ is the sum of $TM_{l}^t$ across the time dimension $t$, denoted as $TM_l$. Therefore, to reduce the number of time steps, we define the mask loss, which is the sum of $TM_{l}$ across all the layers.
However, our experimental results indicate that naively applying this loss to each layer, incurs a large drop in the SNN test accuracy.
To mitigate this issue, we analyze the input distribution of each LIF layer for the spiking ViT model at different time steps. For example, in Fig. \ref{fig:dist:combine}, we visualize the distribution map of some representative spiking ViT layers at the first and the sixth time step with a total of six time steps.

For any multi-time-steps SNN, each LIF layer distribution map can change either slightly or significantly across different time steps. We can interpret the distribution map at the first time step as the baseline, which can change in the subsequent time steps based on the sensitivity (that represents the distribution similarity) of the layer with the different time steps.
For example, if the distribution map of layer $l$ in the first time step is similar to its distribution in all the remaining time steps, it is likely that the layer $l$ can be assigned only a single time step without a significant accuracy drop. Furthermore, considering each distribution map as a probability density vector, we use the cosine similarity metric to quantitatively measure the sensitivity. 
Empirically, we observe that the input distribution of the LIF layers in the MSA, combined across the different time steps, is similar between different blocks. Based on that, we further assume that the sensitivity to the different time steps should be linked to certain groups of layers instead of individual layers. Therefore, we bin all the spiking ViT layers into four categories, the "sps" part, the "qkv" part, the "attn" part, and the "mlp" part, as shown in Fig 2. The "sps" part represents the layers in the Spiking Patch Splitting (SPS) module \cite{spikformer}. The "qkv" part represents the layers that compute the query, key, and value tensors from the inputs. The "attn" part represents the QKV matrix dot-product followed by the scaling operation and the subsequent linear operation after the "qkv" part in the MSA block. The "mlp" part indicates the multi-layer perception block. Each part has varying sensitivites to the different time steps as shown in Fig. 2, which controls the decision to skip a particular time step differently.

\subsection{Training Loss}
Based on the above findings, we compute the loss function of our SNN as the sum of the cross entropy loss (computed from surrogate gradient descent) and the mask loss, as illustrated in Eq.\eqref{eq:loss}, where, $l_s$ denotes the index of the layers that are sufficiently sensitive to the different time steps and the hyperparameter $\lambda_m$ helps to control the accuracy-efficiency trade-off. 
During the surrogate gradient descent, the weights of the SNN are updated based on the gradients to the normal SNNs but with a $TP^t_l$, which is illustrated in Eq.\eqref{eq:bp:weight}. However, computing the gradients of $TP^t_l$ is not straight forward. As illustrated in Eq.\eqref{eq:bp:param}, its gradient flows from both $L_{CE}$ and $L_{mask}$ in two separate directions.

\begin{equation}\label{eq:loss}
    L_{total} = L_{CE} + L_{mask} = L_{CE} + \lambda_m \sum_{l_s}\sum_t {TM}_{l_s}^t
\end{equation}
\vspace{-1mm}
\begin{equation}\label{eq:bp:weight}
    \frac{\partial L_{total}}{\partial W_l} = 
    \frac{\partial L_{CE}}{\partial O_l^{t}} \frac{\partial O_l^{t}}{\partial W_l} = 
    {TM}^{t}_l
    \frac{\partial L_{CE}}{\partial O_l^{t}} \frac{\partial Y_l^{t}}{W_l}
\end{equation}
\begin{equation}\label{eq:bp:param}
    \frac{\partial L_{total}}{\partial TP^{t}_l} =
     = 
    Y_l^t
    \frac{\partial L_{CE}}{\partial O_l^{t}} 
    \frac{\partial TM_l^t}{\partial TP_{l}^t} + 
    \frac{\partial \norm{TM^{t}_l}_1 }{TM^{t}_l}
    \frac{\partial TM^{t}_l}{\partial TP^{t}_l}
\end{equation}
\begin{equation}
    \frac{\partial TM^{t}_l}{\partial TP^{t}_l} = 
    \frac{\partial TM^{t}_l}{\partial TS^{t}_l}
    \frac{\partial TS^{t}_l}{\partial TP^{t}_l}
\end{equation}

\section{Experimental Results}

\subsection{Datasets \& Models}

Our models are based on the compact ViT architecture \cite{compactVit} in which we replace all the linear layers with convolutional layers and remove their bias parameters. 
We performed object recognition experiments on the CIFAR10/100~\cite{cifar10} and ImageNet~\cite{hinton_nips} datasets. The architecture name "Spikformer-N-D" indicates that the model is based on the spikformer architecture \cite{spikformer} and has N self-attention encoder blocks, and the feature embedding dimension is D. For the CIFAR10/100 dataset, we scale the image resolution to $32{\times}32$ and use the spikformer-4-384 architecture. For training, we use Adamw~\cite{adamw} optimizer with an initial learning rate of 0.004, and a batch size of 256. On the other hand, for the ImageNet dataset, we scale the image resolution to $224{\times}224$ and use the spikformer-8-512 architecture to evaluate our approach on large-size models. The initial learning rate is 0.001, the total batch size is 168, and we train for a total of 300 epochs.

\begin{table}
\caption{Comparison of our dynamic time-step spiking ViT models to the original spikformer model with different time steps.$^*$ denotes results based on our replicated setup.}
\label{tab:res:comparsion}
\centering
\renewcommand{\arraystretch}{1.}
\begin{tabular}{l|c|c|c}
\hline
Reference & Architecture & Acc. (\%) & Time steps \\
\hline

\hline
 \multicolumn{4}{|c|}{Dataset : CIFAR10} \\
\hline

CCT\cite{compactVit} & Spikformer-4-384 & 96.61 & 1 \\
\hline
\multirow{4}{*}{Spikformer\cite{spikformer}}  & \multirow{4}{*}{Spikformer-4-384} & 94.13 & 1 \\

&  & 95.29 & 2  \\
&  & 95.66 & 4  \\
&  & 95.94 & 6  \\
\hline
\multirow{2}{*}{\textbf{This work}}  & \multirow{2}{*}{Spikformer-4-384} & \textbf{94.39} & \textbf{1.15}  \\
 & & \textbf{95.97} & \textbf{4.79} \\
\hline
\hline
 \multicolumn{4}{|c|}{Dataset : CIFAR100} \\
\hline
\hline
CCT\cite{compactVit} & Spikformer-4-384 & 80.19 & 1 \\
\hline
\multirow{5}{*}{Spikformer\cite{spikformer}}  & \multirow{5}{*}{Spikformer-4-384} & 75.05 & 1 \\

 & & 77.10 & 2  \\
 & & 77.61 & 3  \\
 & & 78.14 & 4 \\
 & & 78.85 & 6  \\
\hline
\multirow{2}{*}{\textbf{This work}}  & \multirow{2}{*}{Spikformer-4-384} & \textbf{78.09} & \textbf{2.73}  \\

 & & \textbf{78.90} & \textbf{3.81}  \\
\hline
 \multicolumn{4}{|c|}{Dataset : ImageNet} \\
\hline
\hline
\multirow{3}{*}{Spikformer\cite{spikformer}}  & \multirow{3}{*}{Spikformer-8-512} &  63.12* & 1 \\
&  & 66.03*  & 2 \\
&  & 73.38 & 4 \\
\hline
\textbf{This work} & Spikformer-8-512 & \textbf{68.04} & \textbf{1.30} \\
\hline
\end{tabular}
\vspace{-5mm}
\end{table}
\vspace{-3mm}
\subsection{Comparison with SOTA ViT-based SNNs} 

Since our spiking ViT network architecture is based on Spikformer\cite{spikformer}, we exhaustively compare the SNN accuracy obtained by our framework with Spikformer for different number of time steps. We observe that Spikformer incurs a drastic drop in test accuracy as the total number of time steps decreases from $2$ to $1$. Therefore, it is important to yield a SNN that can optimize this trade-off between the test accuracy and time steps. As shown in Table 1, our framework can improve the accuracy by 2.01\% in the ImageNet dataset with a decrease in the total number of time steps by 46\% compared to the one-time-step model. Our SNNs can thus obtain a balanced and controlled (with the hyperparameter $\lambda_m$ shown in Eq. 10) accuracy-efficiency trade-off compared to the spikformer models with both one and two time steps. Furthermore, as also shown in Table 1, for the CIFAR10/100 dataset, our framework can achieve similar accuracy as Spikformer with 6 time steps while reducing the latency ($T_{avg}$) by about {20\%} (from 6 to 4.79) and {36.5\%} (from 6 to 3.81). 
\vspace{-3mm}
\subsection{Comparison with SOTA CNN-based SNNs}

As shown in Table \ref{tab:res:comp:oth}, the SNNs obtained by our framework outperforms the existing CNN-based SNNs in terms of latency-accuracy trade-off. Although, unlike ViTs, the CNN-based SNNs can achieve accuracies similar to the non-spiking CNNs even at low latencies ($1{-}4$ time steps), they are still relatively far from the SOTA that ViTs can provide. Our framework can help achieve this SOTA while still keeping the number of time steps low.

As shown in Table 2, applying the SOTA CNN-based Hoyer regularized training method \cite{datta2022hoyer} (that achieves impressive SNN accuracies with only one time step) on ViTs, lead to a significant accuracy drop in CIFAR10/100. This is because it cannot cleanly separate the input distribution of the LIF layers in the MSA block around the threshold, causing training convergence issues. Other training techniques, including TET \cite{deng2022temporal}, also cannot achieve close to SOTA accuracies, when applied to ViTs.

\begin{table}
\caption{Comparison of our dynamic time-step spiking ViT models to existing CNN and ViT-based low-latency SNNs. $^*$ denotes results are based on our replicated setup.}
\label{tab:res:comp:oth}
\centering
\renewcommand{\arraystretch}{1.}
\begin{tabular}{l|c|c|c}
\hline
Reference & Architecture & Acc. (\%) & Time steps \\
\hline

\hline
 \multicolumn{4}{|c|}{Dataset : CIFAR10} \\
\hline
Diet-SNN\cite{rathi2021diet} & VGG16 & 93.44 & 10 \\
STBP-tdBN\cite{Zheng_Wu_Deng_Hu_Li_2021} & ResNet-19 & 93.16 & 6 \\
Hoyer-reg\cite{datta2022hoyer} & VGG16 & 93.44 & 1 \\
DSR\cite{meng2022training} & PreAct-ResNet-18 & 95.40 & 20 \\
TET\cite{deng2022temporal} & ResNet-19 & 94.44 & 4 \\
\hline
Hoyer-reg\cite{datta2022hoyer} & Spikformer-4-384 & 93.69* & 2 \\
TET\cite{deng2022temporal} & Spikformer-4-384 & 94.08* & 2 \\
\hline
\multirow{2}{*}{\textbf{This work}}  & \multirow{2}{*}{Spikformer-4-384} & \textbf{94.39} & \textbf{1.15} \\

 & & \textbf{95.97} & \textbf{4.79} \\
\hline
\hline
 \multicolumn{4}{|c|}{Dataset : CIFAR100} \\
\hline
\hline
Diet-SNN\cite{rathi2021diet} & VGG16 & 69.67 & 5 \\
STBP-tdBN\cite{Zheng_Wu_Deng_Hu_Li_2021} & ResNet-19 & 70.86 & 4 \\
Hoyer-reg\cite{datta2022hoyer} & Spikformer-4-384 & 71.55 & 1 \\
DSR\cite{meng2022training} & PreAct-ResNet-18 & 78.50 & 20 \\
TET\cite{deng2022temporal} & ResNet-19 & 74.47 & 4 \\
\hline
Hoyer-reg\cite{datta2022hoyer} & Spikformer-4-384 & 71.29* & 2 \\
TET\cite{deng2022temporal} & Spikformer-4-384 & 74.88* & 2 \\
\hline
\multirow{2}{*}{\textbf{This work}}  & \multirow{2}{*}{Spikformer-4-384} & \textbf{78.09} & \textbf{2.73} \\
 & & \textbf{78.90} & \textbf{3.81} \\
\hline
\end{tabular}
\vspace{-3mm}
\end{table}

\vspace{-1mm}

\subsection{Ablation study} 

In this subsection, we demonstrate the impact of mask loss on the accuracy and the FLOPs of our SNNs through two groups of controlled experiments. One group is related to the count of different components in the mask loss, while the other is related to the weight of the mask loss.

The results in Table \ref{tab:ablation:loss} validate our hypothesis described in the {Section \ref{sec:dist}}. 
All the experiments are trained and tested on the CIFAR100 dataset. The initial time step is $2$, and the weight of the mask loss is $1e{-}6$. 
As shown in the first row of Table 3, naively incorporating the mask loss for all the layers can significantly reduce the total number of time steps and hence, the FLOPs, but the accuracy also suffers significantly; the last row indicates that, without the regularization provided by the mask loss, the model still has a large number of time steps after training. Considering the specificity of the SPS module, when we remove the "sps loss" (see second row), the accuracy remains similar, however, the total number of time steps reduces by a factor of $2$. Furthermore, based on our analysis of the sensitivity of layers to the different time steps, we observe that omitting the mask loss of the self-attention layers (along with SPS layers) yields the highest accuracy (see third row), and the number of time steps is reasonably small. 

\begin{table}[t]
\caption{Ablation study results on different mask loss combinations, where TS denotes the total number of time steps, and SA denotes spiking activity.}
\label{tab:ablation:loss}
\centering
\renewcommand{\arraystretch}{1.}
\begin{tabular}{|c|c|c|c|c|c|c|c|}
\hline
\makecell{q\/k\/v\\loss}   &  \makecell{attn\\loss} & \makecell{mlp\\loss} & \makecell{sps\\loss} & Acc. (\%) & TS & SA(\%)  & SOPs (M) \\
\hline
\cmark & \cmark & \cmark & \cmark & 74.64 & 1.03 & 2.90 & 101.24 \\
\cmark & \cmark & \cmark & \xmark & 77.16 & 1.96 & 4.87 & 181.50 \\
\cmark & \xmark & \cmark & \xmark & 78.09 & 2.73 & 5.56 & 185.31 \\
\xmark & \xmark & \xmark & \xmark & 77.10 & 3.93 & 5.33 & 182.69 \\

\hline
\end{tabular}
\vspace{-3mm}
\end{table}

The experiments in Table \ref{tab:ablation:init} use the mask loss consisting of "qkv" loss and "mlp" loss since this configuration provided the best accuracy.
When initialized with a small initial time step, a smaller weight of mask loss $\lambda_m$ incurs significantly higher spiking activity and FLOPs (see first two rows). However, when the initial time step increases, the effect of the $\lambda_m$ becomes less pronounced. Note that a too-small initial time step fails to get high accuracy, and a too-large initial time step fails to reduce the total number of time steps and hence, the FLOPs significantly. When both the initial time step and the maximum time step are equal (see the seventh row), although the total number of time steps do not change after training, the accuracy improves significantly. This indicates that our mask loss may help the model find a better local optimization. We empirically observe initializing the time step to half of the maximum value yields the best accuracy-efficiency trade-off.

\begin{table}[t]
\caption{Ablation study results on the different weights of mask loss and different initial time steps, where $\lambda_m$ denotes the weight of mask loss, TS denotes the number of time steps, and SA denotes spiking activity.}
\label{tab:ablation:init}
\centering
\renewcommand{\arraystretch}{1.}
\begin{tabular}{|c|c|c|c|c|c|c|c|c|}
\hline
\makecell{Init.\\TS} & \makecell{Max.\\TS} & $\lambda_m$ & Acc. (\%) & TS & SA(\%) & SOPs (M)  \\
\hline
1 & 4 & 1e-4 & 74.45 & 2.45 & 3.73 & 172.54 \\
1 & 4 & 1e-6 & 73.90 & 3.0 & 20.97 & 429.18 \\
2 & 4 & 1e-4 & 76.96 & 2.69 & 5.41 & 183.65 \\
2 & 4 & 1e-6 & \textbf{78.09} & \textbf{2.73} & 5.33 & 185.31 \\
3 & 4 & 1e-4 & 78.16 & 3.36 & 7.70 & 254.39 \\
3 & 4 & 1e-6 & 78.07 & 3.36 & 7.71 & 255.78 \\
4 & 4 & 1e-4 & 78.69 & 4.0 & 9.85 & 319.75 \\
4 & 6 & 1e-4 & \textbf{78.90} & 3.81 & 6.54 & 319.51 \\
\hline
\end{tabular}
\vspace{-1mm}
\end{table}

\subsection{Energy Efficiency of our Proposed Approach}

We compare the energy-efficiency of our dynamic time step SNNs with non-spiking ViTs and existing spikformer-based SNNs in Table 5. The compute-efficiency of SNNs stems from two factors: 1) sparsity, that reduces the number of floating point operations (FLOPs) {in convolutional, linear, and MSA layers} compared to non-spiking DNNs according to $SNN^{flops}_l=S_l\times DNN^{flops}_l$, where $S_l$
denotes the average number of spikes per neuron per inference over all timesteps in layer $l$. Note that the sparsity induces a small overhead of checking whether the 1-bit activation is zero, which consumes 0.05pJ in 28nm Kintex-7 FPGA platform according to our post place-and-route simulations. 
2) Use of only AC (1.8pJ) operations that consume $7.4\times$ lower compared to each MAC (13.32pJ) operation in 45nm CMOS technology \cite{horowitz20141} in our FPGA setup for floating-point (FP) representation. The binary activations can replace the FP multiplications with logical operations, i.e., conditional assignment to 0 with a bank of AND gates. 

\begin{table}[t]
\caption{Comparison of the normalized compute (CE), memory (ME), and total (TE) energy between our dynamic time step SNN (DTS SNN), SOTA spikformer-based SNN (Spik. SNN) and non-spiking ViT (NS ViT). The number of time steps (TS) for the dynamic time step and spikformer-based SNN are chosen at iso-accuracy.}
\label{tab:ablation:loss}
\centering
\renewcommand{\arraystretch}{1.}
\begin{tabular}{|c|c|c|c|c|c|}
\hline
Dataset & Type  & TS & \makecell{Norm.\\CE} & \makecell{Norm.\\ME} & \makecell{Norm.\\TE} \\
\hline
CIFAR10 & NS ViT & - & 22.57$\times$ & 0.84$\times$ & 16.12$\times$ \\
CIFAR10 & Spik. SNN & 4 & 2.79$\times$ & 3.68$\times$ & 2.98$\times$ \\
CIFAR10 & DTS SNN & 1.15 & 1$\times$ & 1$\times$ & 1$\times$ \\
\hline
CIFAR100 & NS ViT & -  & 25.7$\times$ & 0.88$\times$ & 17.15$\times$ \\
CIFAR100 & Spik. SNN & 6 & 1.70$\times$ & 2.32$\times$ & 1.89$\times$ \\
CIFAR100 & DTS SNN & 3.81 & 1$\times$ & 1$\times$ & 1$\times$ \\
\hline
\end{tabular}
\vspace{-3mm}
\end{table}

Since our dynamic time step SNNs skip time steps, and consequently, skip more FLOPs compared to existing spikformer-based SNNs (and even more FLOPs compared to non-spiking ViTs that do not have sparsity), as shown in Table 1, our proposed approach leads to higher compute efficiency. These lower FLOPs, coupled with the $7.4\times$ reduction for AC operations leads to a $22.57\times$ and $25.7\times$ reduction in compute energy on CIFAR10 and CIFAR100 respectively compared to non-spiking ViTs with Spikformer-4-384.

Additionally, our SNNs also enjoy superior memory-efficiency compared to existing SNNs since the latter requires less number of membrane potential reads and writes to and from the on-chip memory. We can avoid the potential access for the skipped time step using zero gating logic, that can be supported in neuromorphic chips, such as Loihi. This potential access dominates the memory footprint of the SNNs during inference \cite{datta2022hoyer,snn_evaluation} since it is not influenced by the SNN sparsity since each potential is the sum of a large number of (typically a few 100s) weight-modulated spikes, and so it is almost impossible that all these spike values are zero for the membrane potential to be kept unchanged at a particular time step. Note that the number of weight read and write accesses can be reduced with the spike sparsity, and thus typically do not dominate the memory footprint of the SNN. However, the exact memory savings would depend on the data re-use and underlying hardware. Assuming a simple weight stationary scheme \cite{sze2017efficient}, our SNNs (1.15 time steps) yield $3.68\times$ and $2.32\times$ reduced memory energy compared to Spikformer (4 time steps) at iso-accuracy on CIFAR10 and CIFAR100 respectively. The total (compute+memory) energy savings become $16.12\times$ and $2.98\times$ compared to non-spiking ViT and Spikingformer on CIFAR10. The detailed energy models used to evaluate these numbers are in \cite{Kundu_2021_WACV, ottati2023spike}, while the energy per operation numbers are obtained from our FPGA simulations.

\section{Discussions}

Since the introduction of deep SNNs, many researchers focused on reducing the total number of time steps while not sacrificing the accuracy compared to the non-spiking counterparts. Existing research on low-time-step SNNs have assigned the same number of time steps to all the layers and obtained close to SOTA accuracy. However, we empirically show that this approach is only limited to CNNs (we are the first to the best of our knowledge), and fails when applied to ViTs. We conduct a comprehensive study of the sensitivity of the layers in the model to the number of time steps in order to guide our decision to select the number of time steps for each layer. Based on this study, we propose a training framework with dynamic time step spiking layers as well as a mask loss. The spiking ViT models obtained by our proposed framework outperform the SOTA SNN models with $20\% \sim 30\%$ lower latency with no drop in test accuracy on CIFAR10/100 datasets. 

Our dynamic time step approach is related to the recently proposed work, SEENN \cite{li2023seenn}, that also adjusts the number of time steps dyanmically. However, SEENN conditions the time steps based on the difficulty of the input sample, and not sensitivity of the SNN block, and is limited to CNNs\footnote{The implementation of SEENN is not made public yet for us to evaluate it on ViTs}. We provide insights on how the sensitivity of the activation distributions in ViT can be leveraged to adjust the time steps independent of the sample difficulty. SEENN can potentially be applied to our work to further improve the accuracy-timestep trade-off.

\bibliographystyle{IEEEtran}
\bibliography{IEEEabrv, biblio}

\begin{thebibliography}{10}
\providecommand{\url}[1]{#1}
\csname url@samestyle\endcsname
\providecommand{\newblock}{\relax}
\providecommand{\bibinfo}[2]{#2}
\providecommand{\BIBentrySTDinterwordspacing}{\spaceskip=0pt\relax}
\providecommand{\BIBentryALTinterwordstretchfactor}{4}
\providecommand{\BIBentryALTinterwordspacing}{\spaceskip=\fontdimen2\font plus
\BIBentryALTinterwordstretchfactor\fontdimen3\font minus \fontdimen4\font\relax}
\providecommand{\BIBforeignlanguage}[2]{{%
\expandafter\ifx\csname l@#1\endcsname\relax
\typeout{** WARNING: IEEEtran.bst: No hyphenation pattern has been}%
\typeout{** loaded for the language `#1'. Using the pattern for}%
\typeout{** the default language instead.}%
\else
\language=\csname l@#1\endcsname
\fi
#2}}
\providecommand{\BIBdecl}{\relax}
\BIBdecl

\bibitem{spike_ratecoding}
M.~Pfeiffer \emph{et~al.}, ``Deep learning with spiking neurons: Opportunities and challenges,'' \emph{Frontiers in Neuroscience}, vol.~12, p. 774, 2018.

\bibitem{dsnn_conversion1}
Y.~Cao \emph{et~al.}, ``Spiking deep convolutional neural networks for energy-efficient object recognition,'' \emph{International Journal of Computer Vision}, vol. 113, pp. 54--66, 05 2015.

\bibitem{dsnn_conversion_abhronilfin}
A.~Sengupta \emph{et~al.}, ``Going deeper in spiking neural networks: {VGG} and residual architectures,'' \emph{Frontiers in Neuroscience}, vol.~13, p.~95, 2019.

\bibitem{deng2021optimal}
\BIBentryALTinterwordspacing
S.~Deng and S.~Gu, ``Optimal conversion of conventional artificial neural networks to spiking neural networks,'' in \emph{International Conference on Learning Representations}, 2021. [Online]. Available: \url{https://openreview.net/forum?id=FZ1oTwcXchK}
\BIBentrySTDinterwordspacing

\bibitem{datta2022date}
G.~Datta and P.~A. Beerel, ``Can deep neural networks be converted to ultra low-latency spiking neural networks?'' in \emph{2022 Design, Automation \& Test in Europe Conference \& Exhibition (DATE)}, vol.~1, no.~1, 2022, pp. 718--723.

\bibitem{Wu_Deng_Li_Zhu_Xie_Shi_2019}
Y.~Wu \emph{et~al.}, ``Direct training for spiking neural networks: Faster, larger, better,'' \emph{Proceedings of the AAAI Conference on Artificial Intelligence}, vol.~33, no.~01, pp. 1311--1318, Jul. 2019.

\bibitem{panda_res}
P.~Panda \emph{et~al.}, ``Toward scalable, efficient, and accurate deep spiking neural networks with backward residual connections, stochastic softmax, and hybridization,'' \emph{Frontiers in Neuroscience}, vol.~14, 2020.

\bibitem{datta_ijcnn}
G.~Datta, S.~Kundu, and P.~A. Beerel, ``Training energy-efficient deep spiking neural networks with single-spike hybrid input encoding,'' in \emph{2021 International Joint Conference on Neural Networks (IJCNN)}, vol.~1, no.~1, 2021, pp. 1--8.

\bibitem{datta2022hoyer}
G.~Datta \emph{et~al.}, ``Hoyer regularizer is all you need for ultra low-latency spiking neural networks,'' \emph{arXiv preprint arXiv:2212.10170}, 2022.

\bibitem{rathi2021diet}
N.~Rathi and K.~Roy, ``Diet-snn: A low-latency spiking neural network with direct input encoding and leakage and threshold optimization,'' \emph{IEEE Transactions on Neural Networks and Learning Systems}, vol.~1, no.~1, pp. 1--9, 2021.

\bibitem{Rathi2020Enabling}
\BIBentryALTinterwordspacing
N.~Rathi, G.~Srinivasan, P.~Panda, and K.~Roy, ``Enabling deep spiking neural networks with hybrid conversion and spike timing dependent backpropagation,'' in \emph{International Conference on Learning Representations}, 2020. [Online]. Available: \url{https://openreview.net/forum?id=B1xSperKvH}
\BIBentrySTDinterwordspacing

\bibitem{datta_frontiers}
\BIBentryALTinterwordspacing
G.~Datta, S.~Kundu, A.~R. Jaiswal, and P.~A. Beerel, ``{ACE-SNN}: {Algorithm-Hardware} co-design of energy-efficient \& low-latency deep spiking neural networks for {3D} image recognition,'' \emph{Frontiers in Neuroscience}, vol.~16, 2022. [Online]. Available: \url{https://www.frontiersin.org/articles/10.3389/fnins.2022.815258}
\BIBentrySTDinterwordspacing

\bibitem{datta_lstm}
G.~Datta, H.~Deng, R.~Aviles, Z.~Liu, and P.~A. Beerel, ``Bridging the gap between spiking neural networks \& lstms for latency \& energy efficiency,'' in \emph{2023 IEEE/ACM International Symposium on Low Power Electronics and Design (ISLPED)}, vol.~1, no.~1, 2023, pp. 1--6.

\bibitem{deng2022temporal}
\BIBentryALTinterwordspacing
S.~Deng \emph{et~al.}, ``Temporal efficient training of spiking neural network via gradient re-weighting,'' in \emph{International Conference on Learning Representations}, 2022. [Online]. Available: \url{https://openreview.net/forum?id=_XNtisL32jv}
\BIBentrySTDinterwordspacing

\bibitem{chowdhury20221-step}
S.~S. Chowdhury \emph{et~al.}, ``Towards ultra low latency spiking neural networks for vision and sequential tasks using temporal pruning,'' in \emph{Computer Vision -- ECCV 2022}.\hskip 1em plus 0.5em minus 0.4em\relax Springer Nature Switzerland, 2022, pp. 709--726.

\bibitem{li2023seenn}
Y.~Li \emph{et~al.}, ``Seenn: Towards temporal spiking early-exit neural networks,'' \emph{arXiv preprint arXiv:2304.01230}, 2023.

\bibitem{vgg}
K.~Simonyan and A.~Zisserman, ``Very deep convolutional networks for large-scale image recognition,'' \emph{arXiv preprint arXiv:1409.1556}, 2014.

\bibitem{resnet}
K.~He \emph{et~al.}, ``Deep residual learning for image recognition,'' in \emph{Proceedings of the IEEE conference on computer vision and pattern recognition}, 2016, pp. 770--778.

\bibitem{dosovitskiy2021an}
\BIBentryALTinterwordspacing
A.~Dosovitskiy \emph{et~al.}, ``An image is worth 16x16 words: Transformers for image recognition at scale,'' in \emph{International Conference on Learning Representations}, 2021. [Online]. Available: \url{https://openreview.net/forum?id=YicbFdNTTy}
\BIBentrySTDinterwordspacing

\bibitem{spikformer}
Z.~Zhou \emph{et~al.}, ``Spikformer: When spiking neural network meets transformer,'' \emph{arXiv preprint arXiv:2209.15425}, 2022.

\bibitem{li2022spikeformer}
Y.~Li \emph{et~al.}, ``Spikeformer: A novel architecture for training high-performance low-latency spiking neural network,'' \emph{arXiv preprint arXiv:2211.10686}, 2022.

\bibitem{leefin2020}
C.~Lee \emph{et~al.}, ``Enabling spike-based backpropagation for training deep neural network architectures,'' \emph{Frontiers in Neuroscience}, vol.~14, 2020.

\bibitem{neftci_surg}
E.~O. {Neftci} \emph{et~al.}, ``Surrogate gradient learning in spiking neural networks: Bringing the power of gradient-based optimization to spiking neural networks,'' \emph{IEEE Signal Processing Magazine}, vol.~36, no.~6, pp. 51--63, 2019.

\bibitem{compactVit}
A.~Hassani \emph{et~al.}, ``Escaping the big data paradigm with compact transformers,'' \emph{arXiv preprint arXiv:2104.05704}, 2021.

\bibitem{cifar10}
A.~Krizhevsky, G.~Hinton \emph{et~al.}, ``Learning multiple layers of features from tiny images,'' 2009.

\bibitem{hinton_nips}
A.~Krizhevsky \emph{et~al.}, ``Imagenet classification with deep convolutional neural networks,'' in \emph{Advances in Neural Information Processing Systems 25}, 2012, pp. 1097--1105.

\bibitem{adamw}
I.~Loshchilov and F.~Hutter, ``Decoupled weight decay regularization,'' \emph{arXiv preprint arXiv:1711.05101}, 2017.

\bibitem{Zheng_Wu_Deng_Hu_Li_2021}
H.~Zheng \emph{et~al.}, ``Going deeper with directly-trained larger spiking neural networks,'' \emph{Proceedings of the AAAI Conference on Artificial Intelligence}, vol.~35, no.~12, pp. 11\,062--11\,070, May 2021.

\bibitem{meng2022training}
Q.~Meng \emph{et~al.}, ``Training high-performance low-latency spiking neural networks by differentiation on spike representation,'' in \emph{Proceedings of the IEEE/CVF Conference on Computer Vision and Pattern Recognition}, 2022, pp. 12\,444--12\,453.

\bibitem{horowitz20141}
M.~Horowitz, ``{Computing's} energy problem (and what we can do about it),'' in \emph{2014 IEEE International Solid-State Circuits Conference Digest of Technical Papers (ISSCC)}, 2014, pp. 10--14.

\bibitem{snn_evaluation}
R.~Yin, A.~Moitra, A.~Bhattacharjee, Y.~Kim, and P.~Panda, ``Sata: Sparsity-aware training accelerator for spiking neural networks,'' \emph{IEEE Transactions on Computer-Aided Design of Integrated Circuits and Systems}, vol.~1, no.~1, pp. 1--1, 2022.

\bibitem{sze2017efficient}
V.~Sze, Y.-H. Chen, T.-J. Yang, and J.~Emer, ``Efficient processing of deep neural networks: A tutorial and survey,'' \emph{arXiv preprint arXiv:1703.09039}, 2017.

\bibitem{Kundu_2021_WACV}
S.~Kundu, G.~Datta, M.~Pedram, and P.~A. Beerel, ``Spike-thrift: Towards energy-efficient deep spiking neural networks by limiting spiking activity via attention-guided compression,'' in \emph{Proceedings of the IEEE/CVF Winter Conference on Applications of Computer Vision (WACV)}, January 2021, pp. 3953--3962.

\bibitem{ottati2023spike}
F.~Ottati, C.~Gao, Q.~Chen, G.~Brignone, M.~R. Casu, J.~K. Eshraghian, and L.~Lavagno, ``To spike or not to spike: A digital hardware perspective on deep learning acceleration,'' \emph{arXiv preprint arXiv:2306.15749}, 2023.

\end{thebibliography}


\end{document}